\documentclass[conference]{IEEEtran}
\IEEEoverridecommandlockouts
\usepackage{cite}
\usepackage{amsmath,amssymb,amsfonts}
\usepackage{algorithmic}
\usepackage{graphicx}
\usepackage{textcomp}
\usepackage{xcolor}

\usepackage{amsfonts}
\usepackage{amsmath}
\usepackage{multirow}
\usepackage{array}
\usepackage{booktabs}
\usepackage[switch]{lineno}
\usepackage{algorithm}
\usepackage{CJKutf8}

\def\BibTeX{{\rm B\kern-.05em{\sc i\kern-.025em b}\kern-.08em
    T\kern-.1667em\lower.7ex\hbox{E}\kern-.125emX}}
\begin{document}

\title{CLLD: Contrastive Learning with Label Distance for Text
	Classification\\

\author{\IEEEauthorblockN{Jinhe Lan\textsuperscript{$^{\ast \dagger}$}, Qingyuan Zhan\textsuperscript{$^{\ast}$}, Chenhao Jiang\textsuperscript{$^{\ast}$}, Kunping Yuan, Desheng Wang}
\IEEEauthorblockA{\textit{Alibaba Group }}
\IEEEauthorblockA{\textit{ \{jinhe.lanjh, zhanqingyuan.zqy, jiangchenhao.jch, kunping.ykp, dengsheng.wangds\}@alibaba-inc.com }}
}

\thanks{\textsuperscript{$^{\ast}$}authors contributed equally to this research.
		\textsuperscript{$^{\dagger}$}corresponding author.}

}

\maketitle

\begin{abstract}
Existed pre-trained models have achieved state-of-the-art performance on various text classification tasks. These models have proven to be useful in learning universal language representations. However, the semantic discrepancy between similar texts cannot be effectively distinguished by advanced pre-trained models, which have a great influence on the performance of hard-to-distinguish classes. To address this problem, we propose a novel Contrastive Learning with Label Distance (CLLD) in this work. 
Inspired by recent advances in contrastive learning, we specifically design a classification method with label distance for learning contrastive classes.
CLLD ensures the flexibility within the subtle differences that lead to different label assignments, and generates the distinct representations for each class having similarity simultaneously.
Extensive experiments on public benchmarks and internal datasets demonstrate that our method improves the performance of pre-trained models on classification tasks. Importantly, our experiments suggest that the learned label distance relieve the adversarial nature of inter-classes.

\end{abstract}


\section{Introduction}
Recently, pre-trained language models (PLMs), which learn universal language representations on large corpus, have attracted lots of attention. Bidirectional Encoder Representations from Transformers (BERT) \cite{devlin-etal-2019-bert} and A Robustly Optimized BERT Pretraining Approach (RoBERTa) \cite{liu2019roberta} have been proved to be an effective way to improve various text classification tasks using only the simple classifier of a fully connected layer.

However, the embeddings representations of similar texts with different semantics in PLMs encoding have high similarity. Because of the closer representation caused by the encoding faultiness, classifier tends to be error-prone. As shown in Table \ref{tb1}, it is easy for BERT to make mistakes on such 
samples. To improve the robustness of PLMs, adversarial training is applied in some studies, \cite{Zhu2020FreeLB, jiang-etal-2020-smart} use gradient-based perturbations in the word embeddings during training and \cite{wang-bansal-2018-robust, michel-etal-2019-evaluation} apply high-quality adversarial textual examples. Although these adversarial methods yield promising performance, a key issue is that the model cannot correctly predict label  when the input has small changes. For example, the semantics of verb has changed greatly from \emph{unsatisfying} to \emph{satisfying}, resulting in completely different sentence meanings.  Recent studies \cite{Kaushik2020Learning, gardner-etal-2020-evaluating} adapt contrastive sets to manually disturb the test instances slightly which generate the negative label. \cite{wang2021cline} constructs semantic negative examples unsupervised to improve the robustness under semantically adversarial attacking. In this work, we explore the ways in which the classifier tends to be error-prone owing to the excessive effects of the semantic similarity when the standard of class label assignments disagrees with the semantic similarity. 

In order to address this problem, we propose Contrastive Learning with Label Distance (CLLD). CLLD is a controllable and effective method to utilize negative examples with different labels and employ contrastive learning to distinguish the semantic discrepancy between similar texts. This study provides new insights into the data-driven and supervised label distance, which can affect classes that are difficult to classify. In CLLD, we design a simple multi-task learning framework, which consists of a classification task and a contrastive learning task. The main task learns an ordinary classification that is difficult to distinguish texts with high similarity but different semantics. Meanwhile, the auxiliary task learns distinct representations with different class labels that generally predicted incorrectly in main task. The label distance uses an adjacency matrix to represent the distance between each class, which can be customized to stretch the classes we want to distinguish. The construction of positive and negative samples is mainly based on \cite{gao2021simcse} with only standard dropout used as noise. Integrating label distances makes contrastive learning dynamic and flexible based on different weights between classes 
, which indicates contrastive learning between negative semantic classes.

\begin{table}[t]
	\centering
	\caption{An example of user analysis in E-commerce reviews. And the prediction results are from the BERT (base version with 12 layers).
	} 
	\label{tb1}
	
	\begin{tabular}{lll}
		\hline
		Sentence               & Label    & Predict  \\ \hline
		\begin{CJK*}{UTF8}{gbsn}发货非常快\end{CJK*} & \begin{CJK*}{UTF8}{gbsn}发货快\end{CJK*} & \begin{CJK*}{UTF8}{gbsn}发货快\end{CJK*} \\ \hline
		\begin{CJK*}{UTF8}{gbsn}发货不慢\end{CJK*}   & \begin{CJK*}{UTF8}{gbsn}发货快\end{CJK*} & \begin{CJK*}{UTF8}{gbsn}发货慢\end{CJK*} \\ \hline
		\begin{CJK*}{UTF8}{gbsn}发货到底快不快\end{CJK*}   & \begin{CJK*}{UTF8}{gbsn}发货咨询\end{CJK*} & \begin{CJK*}{UTF8}{gbsn}发货慢\end{CJK*} \\ \hline
	\end{tabular}
\end{table}

To evaluate these proposed methods, extensive experiments are performed on  standard benchmarks of five classification datasets. Results show that CLLD is robust to the subject and achieves the state-of-the-art performance on these datasets. The main contributions of this work are three folds:
\begin{itemize}
	\item A supervised multi-task contrastive learning framework is proposed to learn more generalizable textual representations, and improve the capability of language understanding and classification.
	\item We construct a label distance to dynamically represent the distance of inter-classes, which is integrated in contrastive learning to effectively separate hard-to-distinguish classes.  
	\item Extensive experiments are carried out on five real-world classification datasets and the proposed method achieves superior performance as compared to previous methods.
\end{itemize}

\section{Related Work}

\subsection{Pre-trained Language Models}
The most common unsupervised task in NLP is probabilistic language modeling (LM), which is a classic probabilistic density estimation problem. Recently  pre-trained language
models achieved great success in LM field. Transformer \cite{NIPS2017_3f5ee243} proposed  self-attention
based architecture which soon becomes the backbone of many following LMs. OpenAI GPT \cite{radford2018improving} and BERT \cite{devlin-etal-2019-bert} that pre-trained on a large network with a large amount of unlabeled data use language model fine-tuning in down stream tasks, which has made a breakthrough in several natural language understanding tasks. BERT uses a masked language model to predict words which are randomly masked or replaced to capture a notable
amount of common-sense knowledge. ALBERT \cite{Lan2020ALBERT} presents two parameter-reduction techniques to lower memory consumption and increase the training speed of BERT. In text classification task, despite the significant improvements from pre-trained LMs, the negative semantic of text are easily ignored by the closer embedding of the encoder.  In this paper, we have further explored more correct semantic representation of BERT for text classification. 
\subsection{ Contrastive Learning}
Self-supervised contrastive representation learning \cite{oord2018representation, hjelm2018learning, wu2018unsupervised, chen2020simple, iter-etal-2020-pretraining, ding2021prototypical, chen2020simple,chen2020improved} is the most competitive methods for learning representations without labels. The main idea is to learn a representation by a contrastive loss which pushes apart dissimilar data pairs while pulling together similar pairs. \cite{luo2020capt} encourages the consistency between representations of the original sequence and its corrupted version via unsupervised instance-wise training signals. \cite{wu2020clear} employs multiple sentence-level augmentation strategies in order to learn a noise-invariant sentence representation. \cite{wang2021cline} constructs semantic negative examples unsupervised to improve the robustness under semantically adversarial attacking. One of the major design choices in these work is how to select the positive and negative pairs. The construction strategy of positive and negative examples mainly focus on samples with opposite or similar semantics, while ignoring the negative and positive semantics between classes. The contrastive relationships of inter-classes will be carefully explored in this work.

\begin{figure*}[t]
	\centering
	\includegraphics[width=1.8\columnwidth]{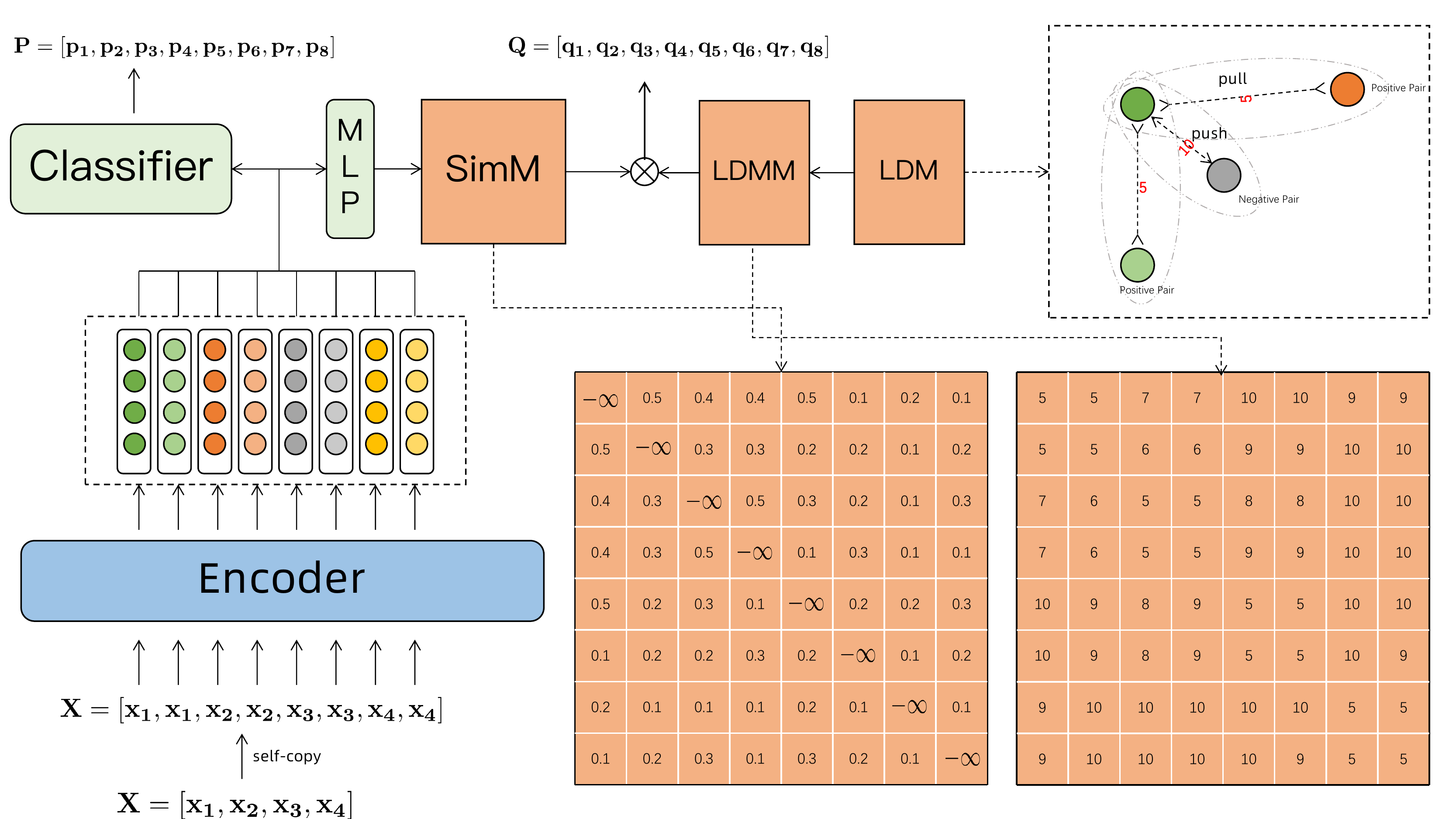} 
	\caption{The framework of CLLD. The main task learns classification, while the auxiliary task gives contrastive supervision with label distance to correct the error-prone class of the main task. SimM denotes the similarity of two samples in a mini-batch (we assume batch size=4). LDMM denotes the distance between the labels of two samples, which is generated by the LDM through vector transformation. In LDM, the value of the matrix denotes the degree of confusion between labels, bringing the samples of the same class closer and the samples of different classes farther.}
	\label{figure1}
\end{figure*}

\section{Contrastive Learning with Label Distance}
In this section, we show the framework of CLLD for text classification, as shown in Fig.\ref{figure1}, which consists of a classification task and a contrastive task. The main task is regarded as a classifier and the auxiliary task is used to help the classifier identify hard-to-distinguish classes more effectively. The neural text encoder $ \mathbb{E}_{\phi}$ parameterized by $\phi$, that maps a sequence of input tokens $\mathbf{x} = [x_1, x_2, \dots, x_T]$ to a sequence of d-dimensional vectors $\mathbf{v}=[v_1, v_2,\dots,v_T]$, $v_{i\in[1:T]}\in\mathbb{R}^d$,  is shared by two tasks and optimized by joint training objective in the training process. In this work, we employ the BERT that has been pre-trained on a large number of corpus as the encoder. 

\begin{equation}
	\mathbf{v}  = \mathbb{E}_{\phi}(\mathbf{x})
	\label{eq0}
\end{equation}

\subsection{Classification Task}
Typical text classification tasks include sentiment analysis, news categorization and topic classification. A dense layer with dropout and a softmax layer are connected in series to form a classifier. The classifier takes an input vector $\mathbf{h}  \in \mathbb{R}^{d}$ from the encoder and produces the predicted probability distribution 
$ \mathbf{p}  \in \mathbb{R}^{N}$ to assign a class label for each sample, where $N$ denotes the number of classes.
\begin{equation}
	\mathbf{p}  = Softmax(\mathbf{W}\mathbf{v} +  \mathbf{b})
	\label{eq1}
\end{equation}
where $\mathbf{W} \in \mathbb{R}^{N \times d}$ and $ \mathbf{b}  \in \mathbb{R}^{N}$
are parameters
of the dense layer and $Softmax(\cdot)$ denotes the softmax function of classification task.

\subsection{Contrastive Task} 
\textbf{Construction of Contrastive Examples.} 
The contrastive learning which aims to learn effective representation
by pulling semantically close neighbors together and pushing apart non-neighbors \cite{hadsell2006dimensionality} is adopted in this work. In order to identify the same and different semantics of examples and add sensitivity to semantic changes to the model, the construction of contrastive examples is particularly important. Gernerally, unsupervised contrastive learning is used to learn more robust text representation before it is employed in downstream tasks.  In our work, we apply this contrastive method directly to the downstream tasks and make model learn the relations of inter-classes. Therefore, samples of the same class are regarded as positives and samples of different classes are regarded as negatives in a mini-batch. 
However,  there may be no similar examples in a batch, resulting in no positive instances. To solve this problem, we simply feed the same input to the encoder twice with only standard dropout used as noise \cite{gao2021simcse}. 
In this way, we take samples of the same class or the vectors disturbed by dropout as positives, so as to increase the diversity and generalization of the construction of the contrastive examples. We employ an Multi-Layer Perceptron (MLP) head upon the base text encoder $\mathbb{E}_{\phi}$ because of contrastive learning can gain better performance when an MLP head is used \cite{He_2020_CVPR},
\begin{equation}
	\mathbf{h}  = MLP(\mathbb{E}_{\phi}(\mathbf{x}))
	\label{eq2}
\end{equation}
where $\mathbf{h}$ denotes the embedding vector generated by the encoder. We assume two set of paired vectors $D = \{(\mathbf{h}_i, \mathbf{h}_i^+), (\mathbf{h}_i, \mathbf{h}_i^-)\}_{i=1}^{n}$ , where $\mathbf{h}_i$ and $\mathbf{h}_i^+$ are positive pairs while $\mathbf{h}_i$ and $\mathbf{h}_i^-$ are negative pairs. $\mathbf{h}_i^+$ can be obtained by sampling from the samples of the same class of $\mathbf{h}_i$ or  through the use of independently sampled dropout masks. $\mathbf{h}_i^-$ is mainly encoded by samples of different classes from $\mathbf{h}_i$. 

\textbf{Similarity Matrix.} 
Having formulated contrastive examples, we now advance to the task of modeling similarity matrix within contrastive learning. To represent the correlation between the contrastive examples, cosine similarity is used to calculate the correlation score between two vectors:
\begin{equation}
sim(\mathbf{h}_i, \mathbf{h}_i^{\{+,-\}}) = {\frac{1}{2}}(\frac{\mathbf{h}_i^{\mathrm{T}} \mathbf{h}_i^{\{+,-\}}}{MAX(\Vert \mathbf{h}_i \Vert_2 \cdot \Vert  \mathbf{h}_i^{\{+,-\}}\Vert_2, \xi ) } + 1)
	\label{eq3}
\end{equation}
where $\Vert \cdot\Vert_2$ denotes 2-norm and $\xi$ is set to small value (we generally use $1e^{-8}$ ) to avoid division by zero. We assume that a mini-batch consists of $M$ examples. For each sample in the batch, we feed sample into the encoder twice, so that our batch size will be increased to $2M$, and then all examples are encoded by the encoder to calculate the pairwise similarity. Through the above method, we can get the similarities between each sample and other samples in a mini-batch, as well as the similarity after dropout disturbance. Due to the same vectors have the highest similarity, self-similarity should be avoided in the similarity matrix by following operation, 
\begin{equation}
	sim(\mathbf{h}_i, \mathbf{h}_i) = -\infty .
	\label{eq4}
\end{equation}

In practice, we set the diagonal elements of the similarity matrix to $-1e^6$, so they have negligible impact in the subsequent calculation. To sum up, we can get the final similarity matrix $\mathbf{SimM}  \in \mathbb{R}^{2M \times 2M}$:

$$
\begin{bmatrix}
	-\infty & sim(\mathbf{h}_1, \mathbf{h}_2)  & ...  & sim(\mathbf{h}_1, \mathbf{h}_{2M}) \\ 
	sim(\mathbf{h}_2, \mathbf{h}_1) & -\infty  & ...  & sim(\mathbf{h}_2, \mathbf{h}_{2M})\\ 
	
	... & ... &  ... & ... \\
	sim(\mathbf{h}_{2M}, \mathbf{h}_1) & sim(\mathbf{h}_{2M}, \mathbf{h}_2) &  ... & -\infty 
\end{bmatrix}
$$

For each example in a mini-batch, the examples of the same class and the duplicate disturbed by dropout are positive instances, and other examples are negative instances. Since we employ a particular operation on the diagonal elements of the similarity matrix, the diagonal elements of the label matrix should be treated particularly for the same purpose. Therefore, the label matrix $[t_{ij}]_{2m\times2m}$  of contrastive task can be calculated by:
\begin{equation}
	t_{ij} = \mathbb{I}(i\neq j)\mathbb{I}(\mathbf{y}_i=\mathbf{y}_j), \ 1\leq i,j\leq2m,
\end{equation}
where $\mathbb{I}(\cdot)$ is an indicator function and $\mathbf{y_i}$ is the one hot vector of the ground truth of the i-th sample. 

\textbf{Label Distance Matrix.} 
The purpose of the contrastive learning is to bring the samples of the same class closer and the samples of different classes farther. However, only considering the same class and different classes ignores the relationship between inter-classes, which leads to the equal treatment of non-same classes and the effect of non-same classes can not be improved by contrastive examples. In order to solve this problem, we propose a simple and efficient method, which constructs a label distance matrix to represent the correlation between different classes. We use the matrix $\mathbf{LDM}=[d_{ij}]_{N\times N}  \in \mathbb{R}^{N \times N} $ to denote label distance whose values can 
denote the correlation between inter-classes. In a mini-batch, we can calculate label distance mask matrix ($\mathbf{LDMM}$) by,

\begin{equation}	
	\small
	\mathbf{LDMM}=\left[\begin{array}{c}
		\mathbf{y}_1 ^{\mathrm{T}}\\
		\vdots \\
		\mathbf{y}_{2M}^{\mathrm{T}}
	\end{array}\right]\left[\begin{array}{ccc}
		d_{11} & \cdots & d_{1,N} \\
		\vdots & \ddots & \vdots \\
		d_{N,1} & \cdots & d_{N,N}
	\end{array}\right]\left[\mathbf{y}_1, \dots,\mathbf{y}_{2M}\right],
\end{equation}
and then
\begin{equation}
	\mathbf{SimLDM} = \mathbf{SimM} \bigotimes \mathbf{LDMM} ,
	\label{eq5}
\end{equation}
where $\bigotimes $ denotes the element-wise multiplication implemented by a broadcast method and $\mathbf{SimLDM}$ denotes the final similarity matrix. The diversity of data and the different classes of each dataset determine that the construcation of label distance is data-driven. In this work, we design static and dynamic construction methods. The performance of static mode is better, but it needs manual design and more troublesome. Algorithm \ref{alg1} provides an adaptive mode for label distance matrix generation and updation. We establish the relationship between confusion matrix and $\mathbf{LDM}$ matrix, so as to distinguish confusing classes. Since the model is not stable at the initial stage of training, $\mathbf{LDM}$ is difficult to build based on the results of evaluation instability. $\mathbf{LDM}$ will not be updated until the model is relatively stable. In our experiment, we start to update $\mathbf{LDM}$  after 10 epoches,  and update $\mathbf{LDM}$ according to the calculated updated intensity $\epsilon$ every ten evaluations. The updated intensity decreases linearly in training process.
\begin{algorithm}
	\renewcommand{\algorithmicrequire}{\textbf{Input:}}
	\renewcommand{\algorithmicensure}{\textbf{Output:}}
	\caption{The Construction of Adaptive Label Distance Matrix}
	\label{alg1}
	\begin{algorithmic}[1]
		\STATE Initialization:$\mathbf{LDM},num\_epoch = n,eval\_steps=k$
		\STATE $\mathbf{LDM} = \{ d_{i,j} | 1 <= i, j <= 2M \}$
		\FOR {$epoch = 1$ to $num\_epoch$}        
		\IF {$epoch > n / 5$ and $training\_steps\ \%\ (10*k)==0$ }
		
		\STATE $\mathbf{p}_{validation} =  MODEL(\mathbf{x}_{validation}, \mathbf{LDM})$
		\STATE $\mathbf{CM}_{N \times N} = Confusion Matrix(\mathbf{y}, \mathbf{p}_{validation})$
		\STATE $\mathbf{LDM}^{'}_{N \times N} = COUNT(\mathbf{CM})$
		\STATE Matrix $\mathbf{LDM}$ is a record of the number of classes predicted by the model for each label based on confusion matrix $\mathbf{CM}$ statistics
		\STATE $\epsilon = 1 - training\_steps/total\_steps$
		\STATE $\mathbf{LDM} = (\mathbf{LDM}^{'} - \mathbf{LDM} )*\epsilon + \mathbf{LDM}$
		\ELSE 
		\STATE NORMAL TRANING
		\ENDIF 
		
		\ENDFOR
	\end{algorithmic}  
\end{algorithm}

\subsection{Training Objectives}
The overall loss function $ \mathcal{L} $ sums the loss of the classification task $\mathcal{L}_c$ and that of the contrastive task $\mathcal{L}_s$:

\begin{equation}
	\mathcal{L}  = \lambda \mathcal{L}_c + (1-\lambda )\mathcal{L}_s,
	\label{eq6}
\end{equation}
where $\lambda $ is used as the hyper parameter of the multi-task training loss to control the intensity of the contrast task.

For classification task, the conventional cross entropy loss is employed to be objective function:
\begin{equation}
	\mathcal{L}_c = - \frac{1}{M}\sum_{i=1}^{M}\sum_{c=1}^{N}\mathbf{y}_{ic}log(\mathbf{p}_{ic}).
	\label{eq7}
\end{equation}

For contrastive task, we construct KL-divergence loss to solve the problem that cross entropy cannot deal with the non-unique positive examples. For an example in a mini-batch, the number of positive examples is non-unique, which should be closer to all positive examples. Therefore, the KL-divergence loss can make the similarity between positive example pairs greater and negative example pairs smaller. The specific formula is as follows:
\begin{equation}
	\mathcal{L}_s = - \frac{1}{2M}\sum_{i=1}^{2M}(\sum_{j=1}^{2M}(\mathbf{z}_{ij}log(\mathbf{q}_{ij})-\mathbf{z}_{ij}log(\mathbf{z}_{ij})))
	\label{eq8}
\end{equation}
where $\mathbf{q}_{ij}$ and $\mathbf{z}_{ij}$:

\begin{equation}
	\mathbf{q}_{ij} = \frac{e^{SimLDM(\mathbf{h}_{i},\mathbf{h}_{j}) }}{\sum_{j=1}^{2M}e^{SimLDM(\mathbf{h}_{i},\mathbf{h}_{j}) }}
	\label{eq9}
\end{equation}

\begin{equation}
\mathbf{z}_{ij} = \frac{t_{ij}}{\sum_{j=1}^{2M}{t_{ij} }}
	\label{eq10}
\end{equation}
 Finally, we get the training objective loss $\mathcal{L}$ of multi-task learning, which will be applied to our experiments.
\section{Experiments}
\subsection{Datasets}
We run our experiments on five widely used datasets including R8 and R52 of Reuters 21578, TREC-6, Movie Review and 20 Newsgroups. We collect user
reviews from Taobao to evaluate the effectiveness of our approach in industrial applications.


\textbf{20 Newsgroups.} The 20 Newsgroups (20NG) dataset\footnote{http://qwone.com/˜jason/20Newsgroups/} contains 18846 documents evenly categorized into 20 different categories. In total, 11,314 documents are in the training set and 7,532 documents are in the test set.


\textbf{R8 \& R52.} R8 and R52 are two widely used subsets of the Reuters-21578 dataset\footnote{https://martin-thoma.com/nlp-reuters.}. R8 consists of samples in 8 categories while R52 in 52 categories. R8 was split to 5,485 training and 2,189 test samples and R52 was split to 6,532 training and 2,568 test samples.

\textbf{Movie Review.} The Movie Review (MR) dataset  is a movie review dataset for binary sentiment classification. It includes 10,662 sentences with even numbers of negative and positive samples. We used the training/test split in \cite{tang2015pte}.


\textbf{TREC-6 }  The TREC dataset\cite{voorhees1999trec} is one of the most popular datasets for question classification. This dataset is known as TREC-6 which consists of questions in 6 categories. The training and test datasets contain 5,452 and 500 questions, respectively.
\begin{table*}[t]
	\renewcommand\arraystretch{1.2}
	\centering
	\caption{Performance comparisons on different datasets.} 
	\label{tb2}
	\begin{tabular}{c|p{1.2cm}<{\centering}p{1.6cm}<{\centering}p{1.1cm}<{\centering}p{1.7cm}<{\centering}p{1.4cm}<{\centering}p{1.8cm}<{\centering}}
		
		\specialrule{0.1em}{0pt}{2pt}
		Datasets        & BERT  & BERT+CLLD & ALBERT & ALBERT+CLLD & TextCNN & TextCNN+CLLD   \\ 
		\specialrule{0.05em}{2pt}{2pt}
		20NG                    & 85.3 &  \textbf{86.6} & 80.6 & \textbf{81.7}  & 78.8 &\textbf{80.0} \\ 
		TREC\_6             & 97.6 & \textbf{97.8}  & 96.4 & \textbf{96.6} & 88.2 & \textbf{88.8}\\ 
		R8     & 97.8 & \textbf{98.1} & 97.3 & \textbf{97.7}  & 94.8 & \textbf{95.0}\\ 
		R52      & 96.4  & \textbf{96.5} & 94.8 & \textbf{95.6} & 91.1 & \textbf{91.3} \\ 
		MR & 85.7  & \textbf{87.0} & 86.6 & \textbf{97.0 }& 70.9 & \textbf{72.1}\\ 
		\specialrule{0.1em}{2pt}{0pt}
	\end{tabular}
	
\end{table*}
\subsection{Experiment Settings}
We use a pre-trained BERT-base (12 layers, 768 hidden vector size, 12 attention heads, 110M parameters) from HuggingFace’s Transformers\cite{wolf2019huggingface} with a dense layer as our base model. We fine-tune the BERT model for a maximum of 50 epoches with a batch size of 128 sequences of maximum length 128 token on 4 GTX1080Ti GPUs and stop if the validation performance does not improve for 10 consecutive epoches. The dropout probability are kept at 0.1 on all layers and in attention in BERT and 0.5 on the dense layer. We use $\tau =5.0$ for the fixed label distance matrix and $\lambda=0.5$ for loss allocation in practice. The base learning rate is 2e-5.  For fine-tuning BERT, we use AdamW as our optimizer with $\beta$ hyperparameters ($\beta_1= 0.9$, $\beta_2 = 0.999$) and a decoupled weight decay \cite{loshchilov2017decoupled} of 0.1. For baseline models, we use default parameter settings as in their original papers or implementations. All of our experimental choices were tuned by observing the performance on the validation sets.

\subsection{Experiment Results}

Table \ref{tb2} compares the performance of CLLD and pre-trained models for text classification on six real-world datasets.
Accuracy is set as the evaluation metric in this work. In order to prove the generality of our method, TextCNN \cite{kim-2014-convolutional} is also used as a baseline. Clearly, our method outperforms all baseline models, which verifies the superiority of CLLD. Moreover, it has achieved a 15\% improvement in the real Taobao review dataset.

It is difficult to distinguish similar texts with negative semantic relationship or different labels only using BERT as encoder to build classification model. Therefore, datasets with more samples and more labels are more likely to have similar texts but different semantics. As is shown table \ref{tb2}, CLLD has the most significant effect on 20NG with the largest number of samples. Since the adaptive update of label distance needs to evaluate the effect of the model on the validation set,the number of samples in the validation set will affect the effect of label distance on the similarity matrix. R8 and TREC, when the number of training set samples is close to the same, the R8 improvement with more verification set samples is greater than that of TREC. 
\subsection{Ablation Studies}
Here, we explore the effectiveness of different task and mechanism of CLLD and we perform an ablation study on 20NG. We compare four baselines, which are with different methods to build the our model. The four baselines are as follows: 
\begin{enumerate}
	\item CT: For classification task(CT), we directly use BERT as the encoder to build the classification model.
	\item CT+CL: On the basis of classification tasks, contrastive learning(CL) tasks are added to build a multi-task learning model. 
	\item CT+CLSLD: Adding contrastive learning with static label distance(CLSLD) to multi-task learning framework. We build static label distance by manually selecting confusing classes in the dataset and manually increasing the distance between these classes.
	\item CT+CLALD:Adding contrastive learning with adaptive label distance(CLALD) to multi-task learning framework.
\end{enumerate}
\begin{table}[t]
	\renewcommand\arraystretch{1.2}
	\centering
	\caption{Ablation studies on 20NG dataset.} 
	\label{tb3}
	\begin{tabular}{c|p{1.6cm}<{\centering}}
		
		\specialrule{0.1em}{0pt}{2pt}
		Methods        & 20NG  \\ 
		\specialrule{0.05em}{2pt}{2pt}
		CT                    & 85.30   \\ 
		CT+CL           & 86.30   \\ 
		CT+CLSLD             & \textbf{86.60}   \\ 
		CT+CLALD     & 86.56   \\ 
		\specialrule{0.1em}{2pt}{0pt}
	\end{tabular}
	
\end{table}
The prediction performances of ablative variants of CLLD are shown in Table \ref{tb3}. As can be seen from the table, multi-task learning framework achieves better result than the previous model, proving the advantages of our multi-task method in describing negative semantic features between samples.  Because label distance not only considers the contrastive samples in the contrastive task, but also realizes the contrastive learning between classes through the distance of inter-classes, which can achieve better performance. Due to the addition of artificial prior-knowledge, the effect of CLSLD is improved more significantly than that of CLALD.

\subsection{Case Study}
\begin{figure}[htb]
	\includegraphics[width=1.5\columnwidth]{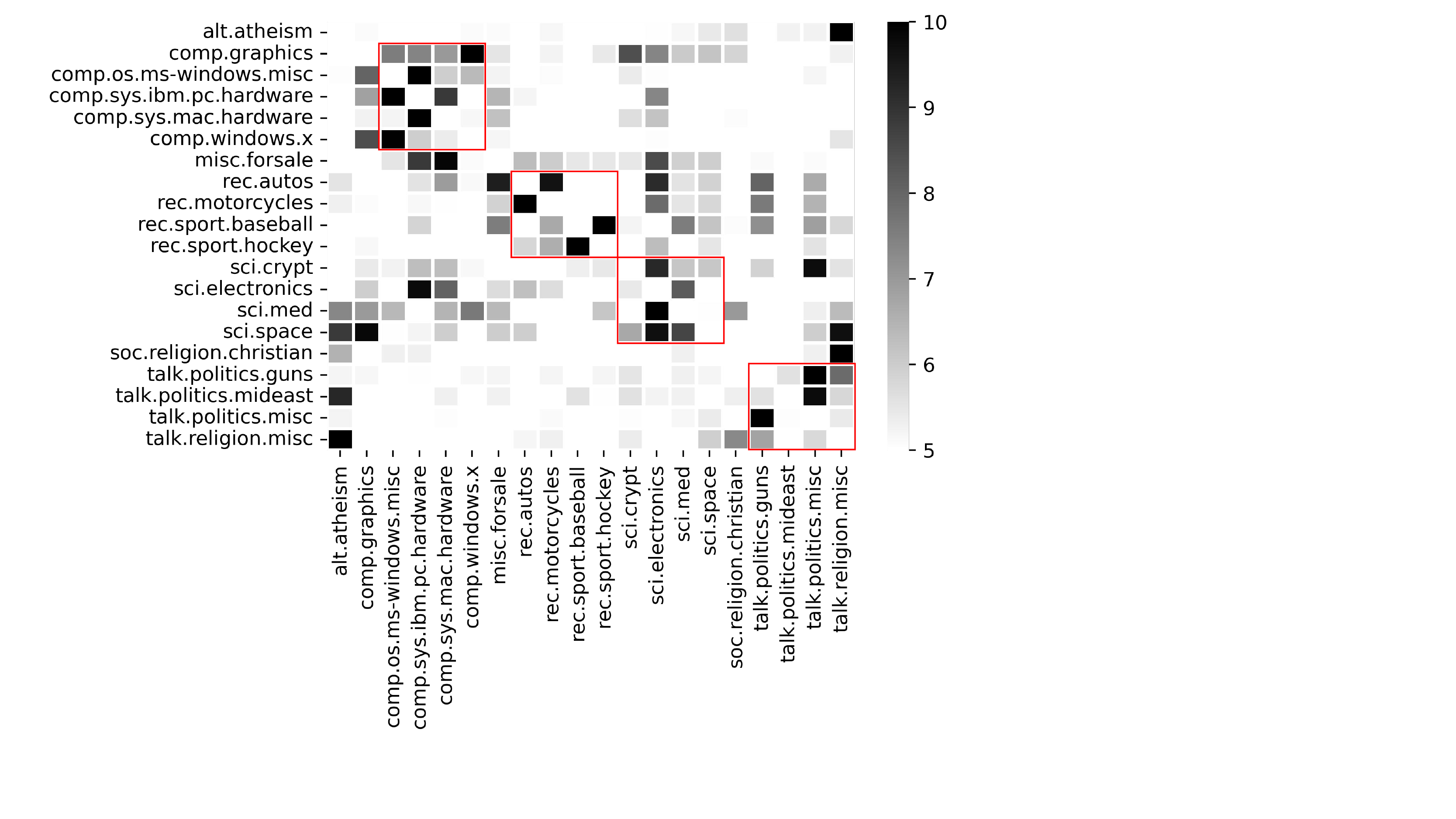} 
	\caption{The matrix of label distance on 20NG dataset. }
	\label{figure2}
\end{figure}

Fig. \ref{figure2} shows an example of the label distance generated by algorithm \ref{alg1} on 20NG dataset. The gray level of each element in the graph indicates the size of the distance. We can see that the \emph{alt.atheism} is further away from the \emph{talk.religion.misc}, which indicates that these two religion-related labels are more likely to be confused by the model. Similarly, the labels framed with red lines in the Fig. \ref{figure2} belong to the same category of labels, and the texts of these labels are similar in syntax. This can be seen that the more hard-to-distinguish the classes, the greater the distance of inter-classes. As this case very clearly demonstrates, it is important that the matrix of label distance have learned the correlation between labels to effectively separate hard-to-distinguish examples.

\section{Conclusion}
In this work, we present a novel contrastive learning with label distance that use contrastive examples with different labels to learn the contrastive relations of inter-classes. The multi-task learning framework brings base pre-trained models the ability to capture the semantic discrepancy between similar texts and effectively distinguish the classes with similar semantics. Extensive experiments demonstrate that the performance of the pre-trained model with CLLD exceeds previous methods. 

In future work, we will examine the construction of label distance in the contrastive task.
Moreover, we will adapt our model to other natural language processing tasks. We expect that our model will encourage
a pre-trained model to generate distinct textual representations with different
classes. 

\section*{Acknowledgment}
This work was supported by Alibaba Group. We would like to thank colleagues of our team for discussion and providing useful feedback on the project.

\bibliographystyle{./IEEEtran}
\bibliography{./mybib}

\begin{thebibliography}{10}
\providecommand{\url}[1]{#1}
\csname url@samestyle\endcsname
\providecommand{\newblock}{\relax}
\providecommand{\bibinfo}[2]{#2}
\providecommand{\BIBentrySTDinterwordspacing}{\spaceskip=0pt\relax}
\providecommand{\BIBentryALTinterwordstretchfactor}{4}
\providecommand{\BIBentryALTinterwordspacing}{\spaceskip=\fontdimen2\font plus
\BIBentryALTinterwordstretchfactor\fontdimen3\font minus
  \fontdimen4\font\relax}
\providecommand{\BIBforeignlanguage}[2]{{%
\expandafter\ifx\csname l@#1\endcsname\relax
\typeout{** WARNING: IEEEtran.bst: No hyphenation pattern has been}%
\typeout{** loaded for the language `#1'. Using the pattern for}%
\typeout{** the default language instead.}%
\else
\language=\csname l@#1\endcsname
\fi
#2}}
\providecommand{\BIBdecl}{\relax}
\BIBdecl

\bibitem{devlin-etal-2019-bert}
J.~Devlin, M.-W. Chang, K.~Lee, and K.~Toutanova, ``Bert: Pre-training of deep
  bidirectional transformers for language understanding,'' in \emph{Proceedings
  of the 2019 Conference of the North {A}merican Chapter of the Association for
  Computational Linguistics: Human Language Technologies, Volume 1 (Long and
  Short Papers)}.\hskip 1em plus 0.5em minus 0.4em\relax Association for
  Computational Linguistics, jun 2019, pp. 4171--4186.

\bibitem{liu2019roberta}
Y.~Liu, M.~Ott, N.~Goyal, J.~Du, M.~Joshi, D.~Chen, O.~Levy, M.~Lewis,
  L.~Zettlemoyer, and V.~Stoyanov, ``Roberta: A robustly optimized bert
  pretraining approach,'' \emph{arXiv preprint arXiv:1907.11692}, 2019.

\bibitem{Zhu2020FreeLB}
C.~Zhu, Y.~Cheng, Z.~Gan, S.~Sun, T.~Goldstein, and J.~Liu, ``Freelb: Enhanced
  adversarial training for natural language understanding,'' in
  \emph{International Conference on Learning Representations}, 2020.

\bibitem{jiang-etal-2020-smart}
H.~Jiang, P.~He, W.~Chen, X.~Liu, J.~Gao, and T.~Zhao, ``{SMART}: Robust and
  efficient fine-tuning for pre-trained natural language models through
  principled regularized optimization,'' in \emph{Proceedings of the 58th
  Annual Meeting of the Association for Computational Linguistics}.\hskip 1em
  plus 0.5em minus 0.4em\relax Online: Association for Computational
  Linguistics, Jul. 2020, pp. 2177--2190.

\bibitem{wang-bansal-2018-robust}
Y.~Wang and M.~Bansal, ``Robust machine comprehension models via adversarial
  training,'' in \emph{Proceedings of the 2018 Conference of the North
  {A}merican Chapter of the Association for Computational Linguistics: Human
  Language Technologies, Volume 2 (Short Papers)}.\hskip 1em plus 0.5em minus
  0.4em\relax New Orleans, Louisiana: Association for Computational
  Linguistics, Jun. 2018, pp. 575--581.

\bibitem{michel-etal-2019-evaluation}
P.~Michel, X.~Li, G.~Neubig, and J.~Pino, ``On evaluation of adversarial
  perturbations for sequence-to-sequence models,'' in \emph{Proceedings of the
  2019 Conference of the North {A}merican Chapter of the Association for
  Computational Linguistics: Human Language Technologies, Volume 1 (Long and
  Short Papers)}.\hskip 1em plus 0.5em minus 0.4em\relax Minneapolis,
  Minnesota: Association for Computational Linguistics, Jun. 2019, pp.
  3103--3114.

\bibitem{Kaushik2020Learning}
D.~Kaushik, E.~Hovy, and Z.~Lipton, ``Learning the difference that makes a
  difference with counterfactually-augmented data,'' in \emph{International
  Conference on Learning Representations}, 2020.

\bibitem{gardner-etal-2020-evaluating}
M.~Gardner, Y.~Artzi, V.~Basmov, J.~Berant, B.~Bogin, S.~Chen, P.~Dasigi,
  D.~Dua, Y.~Elazar, A.~Gottumukkala, N.~Gupta, H.~Hajishirzi, G.~Ilharco,
  D.~Khashabi, K.~Lin, J.~Liu, N.~F. Liu, P.~Mulcaire, Q.~Ning, S.~Singh, N.~A.
  Smith, S.~Subramanian, R.~Tsarfaty, E.~Wallace, A.~Zhang, and B.~Zhou,
  ``Evaluating models{'} local decision boundaries via contrast sets,'' in
  \emph{Findings of the Association for Computational Linguistics: EMNLP
  2020}.\hskip 1em plus 0.5em minus 0.4em\relax Online: Association for
  Computational Linguistics, Nov. 2020, pp. 1307--1323.

\bibitem{wang2021cline}
D.~Wang, N.~Ding, P.~Li, and H.-T. Zheng, ``Cline: Contrastive learning with
  semantic negative examples for natural language understanding,'' \emph{Annual
  Meeting of the Association for Computational Linguistics}, 2021.

\bibitem{gao2021simcse}
T.~Gao, X.~Yao, and D.~Chen, ``{SimCSE}: Simple contrastive learning of
  sentence embeddings,'' 2021.

\bibitem{NIPS2017_3f5ee243}
A.~Vaswani, N.~Shazeer, N.~Parmar, J.~Uszkoreit, L.~Jones, A.~N. Gomez, L.~u.
  Kaiser, and I.~Polosukhin, ``Attention is all you need,'' in \emph{Advances
  in Neural Information Processing Systems}, I.~Guyon, U.~V. Luxburg,
  S.~Bengio, H.~Wallach, R.~Fergus, S.~Vishwanathan, and R.~Garnett, Eds.,
  vol.~30.\hskip 1em plus 0.5em minus 0.4em\relax Curran Associates, Inc.,
  2017.

\bibitem{radford2018improving}
A.~Radford, K.~Narasimhan, T.~Salimans, and I.~Sutskever, ``Improving language
  understanding by generative pre-training (2018),'' 2018.

\bibitem{Lan2020ALBERT}
\BIBentryALTinterwordspacing
Z.~Lan, M.~Chen, S.~Goodman, K.~Gimpel, P.~Sharma, and R.~Soricut, ``Albert: A
  lite bert for self-supervised learning of language representations,'' in
  \emph{International Conference on Learning Representations}, 2020. [Online].
  Available: \url{https://openreview.net/forum?id=H1eA7AEtvS}
\BIBentrySTDinterwordspacing

\bibitem{oord2018representation}
A.~v.~d. Oord, Y.~Li, and O.~Vinyals, ``Representation learning with
  contrastive predictive coding,'' \emph{arXiv preprint arXiv:1807.03748},
  2018.

\bibitem{hjelm2018learning}
R.~D. Hjelm, A.~Fedorov, S.~Lavoie-Marchildon, K.~Grewal, P.~Bachman,
  A.~Trischler, and Y.~Bengio, ``Learning deep representations by mutual
  information estimation and maximization,'' in \emph{International Conference
  on Learning Representations}, 2019.

\bibitem{wu2018unsupervised}
Z.~Wu, Y.~Xiong, S.~X. Yu, and D.~Lin, ``Unsupervised feature learning via
  non-parametric instance discrimination,'' in \emph{Proceedings of the IEEE
  conference on computer vision and pattern recognition}, 2018, pp. 3733--3742.

\bibitem{chen2020simple}
T.~Chen, S.~Kornblith, M.~Norouzi, and G.~Hinton, ``A simple framework for
  contrastive learning of visual representations,'' in \emph{International
  conference on machine learning}.\hskip 1em plus 0.5em minus 0.4em\relax PMLR,
  2020, pp. 1597--1607.

\bibitem{iter-etal-2020-pretraining}
D.~Iter, K.~Guu, L.~Lansing, and D.~Jurafsky, ``Pretraining with contrastive
  sentence objectives improves discourse performance of language models,'' in
  \emph{Proceedings of the 58th Annual Meeting of the Association for
  Computational Linguistics}.\hskip 1em plus 0.5em minus 0.4em\relax Online:
  Association for Computational Linguistics, Jul. 2020, pp. 4859--4870.

\bibitem{ding2021prototypical}
N.~Ding, X.~Wang, Y.~Fu, G.~Xu, R.~Wang, P.~Xie, Y.~Shen, F.~Huang, H.-T.
  Zheng, and R.~Zhang, ``Prototypical representation learning for relation
  extraction,'' in \emph{International Conference on Learning Representations},
  2021.

\bibitem{chen2020improved}
X.~Chen, H.~Fan, R.~Girshick, and K.~He, ``Improved baselines with momentum
  contrastive learning,'' \emph{arXiv preprint arXiv:2003.04297}, 2020.

\bibitem{luo2020capt}
F.~Luo, P.~Yang, S.~Li, X.~Ren, and X.~Sun, ``Capt: Contrastive pre-training
  for learning denoised sequence representations,'' \emph{arXiv preprint
  arXiv:2010.06351}, 2020.

\bibitem{wu2020clear}
Z.~Wu, S.~Wang, J.~Gu, M.~Khabsa, F.~Sun, and H.~Ma, ``Clear: Contrastive
  learning for sentence representation,'' \emph{arXiv preprint
  arXiv:2012.15466}, 2020.

\bibitem{hadsell2006dimensionality}
R.~Hadsell, S.~Chopra, and Y.~LeCun, ``Dimensionality reduction by learning an
  invariant mapping,'' in \emph{2006 IEEE Computer Society Conference on
  Computer Vision and Pattern Recognition (CVPR'06)}, vol.~2.\hskip 1em plus
  0.5em minus 0.4em\relax IEEE, 2006, pp. 1735--1742.

\bibitem{He_2020_CVPR}
K.~He, H.~Fan, Y.~Wu, S.~Xie, and R.~Girshick, ``Momentum contrast for
  unsupervised visual representation learning,'' in \emph{Proceedings of the
  IEEE/CVF Conference on Computer Vision and Pattern Recognition (CVPR)}, June
  2020.

\bibitem{tang2015pte}
J.~Tang, M.~Qu, and Q.~Mei, ``Pte: Predictive text embedding through
  large-scale heterogeneous text networks,'' in \emph{Proceedings of the 21th
  ACM SIGKDD international conference on knowledge discovery and data mining},
  2015, pp. 1165--1174.

\bibitem{voorhees1999trec}
E.~M. Voorhees \emph{et~al.}, ``The trec-8 question answering track
  report.''\hskip 1em plus 0.5em minus 0.4em\relax Citeseer.

\bibitem{wolf2019huggingface}
T.~Wolf, L.~Debut, V.~Sanh, J.~Chaumond, C.~Delangue, A.~Moi, P.~Cistac,
  T.~Rault, R.~Louf, M.~Funtowicz \emph{et~al.}, ``Huggingface's transformers:
  State-of-the-art natural language processing,'' \emph{arXiv preprint
  arXiv:1910.03771}, 2019.

\bibitem{loshchilov2017decoupled}
I.~Loshchilov and F.~Hutter, ``Decoupled weight decay regularization,''
  \emph{arXiv preprint arXiv:1711.05101}, 2017.

\bibitem{kim-2014-convolutional}
Y.~Kim, ``Convolutional neural networks for sentence classification,'' in
  \emph{Proceedings of the 2014 Conference on Empirical Methods in Natural
  Language Processing ({EMNLP})}.\hskip 1em plus 0.5em minus 0.4em\relax Doha,
  Qatar: Association for Computational Linguistics, Oct. 2014, pp. 1746--1751.

\end{thebibliography}
\end{document}